\documentclass[12pt,authoryear,preprint]{elsarticle}

\usepackage{array}
\usepackage{booktabs}
\usepackage{multirow}
\usepackage[figuresright]{rotating}
\usepackage{oubraces}
\newcommand{\vc}[1]{\boldsymbol{\mathbf{#1}}}
\usepackage{amsmath}
\usepackage{mathtools}
\usepackage{subfigure}
\usepackage{algorithm}
\usepackage{algorithmic}
\usepackage{adjustbox}
\usepackage{newfloat}
\usepackage{listings}

\usepackage{color}
\usepackage{amsmath}
\usepackage{amssymb}
\usepackage{bm}
\usepackage{url}
\usepackage[colorlinks=true]{hyperref}
\usepackage{multirow}
\usepackage{booktabs}
\usepackage{multirow}
\usepackage[figuresright]{rotating}
\usepackage{oubraces}
\usepackage{mathtools}
\usepackage{subfigure}
\usepackage{rotating}
\usepackage{tabu}
\DeclareMathSymbol{\R}{\mathalpha}{AMSb}{"52}

\definecolor{bostonuniversityred}{rgb}{0.8, 0.0, 0.0}

\journal{Neural Networks}

\begin{document}

\begin{frontmatter}
\title{Drop Edges and Adapt: a Fairness Enforcing Fine-tuning for Graph Neural Networks}

\author[sapienza]{Indro Spinelli}

\author[]{Riccardo Bianchini}

\author[sapienza]{Simone Scardapane\corref{cor1}}
\ead{simone.scardapane@uniroma1.it}
\cortext[cor1]{Corresponding author. Phone: +39 06 44585495, Fax: +39 06 4873300.}
\address[sapienza]{Department of Information Engineering, Electronics and Telecommunications (DIET), Sapienza University of Rome, Via Eudossiana 18, 00184 Rome, Italy}



\begin{abstract}
The rise of graph representation learning as the primary solution for many different network science tasks led to a surge of interest in the fairness of this family of methods. Link prediction, in particular, has a substantial social impact. However, link prediction algorithms tend to increase the segregation in social networks by disfavoring the links between individuals in specific demographic groups. This paper proposes a novel way to enforce fairness on graph neural networks with a fine-tuning strategy. We \textbf{D}rop the unfair \textbf{E}dges and, simultaneously, we \textbf{A}dapt the model's parameters to those modifications, DEA in short. We introduce two covariance-based constraints designed explicitly for the link prediction task. We use these constraints to guide the optimization process responsible for learning the new `fair' adjacency matrix. One novelty of DEA is that we can use a discrete yet learnable adjacency matrix in our fine-tuning. We demonstrate the effectiveness of our approach on five real-world datasets and show that we can improve both the accuracy and the fairness of the link prediction tasks. In addition, we present an in-depth ablation study demonstrating that our training algorithm for the adjacency matrix can be used to improve link prediction performances during training. Finally, we compute the relevance of each component of our framework to show that the combination of both the constraints and the training of the adjacency matrix leads to optimal performances.

\end{abstract}


\end{frontmatter}

\section{Introduction}
\label{sec:introduction}

The fairness of graph representation learning algorithms is quickly becoming a crucial area of research. Of particular interest is the fairness issue associated with the link prediction task. This task is heavily applied in two of the most influential AI-powered domains of our digital life, social networks and products recommendation. Social network topologies define the stream of information we will receive, often influencing our opinion \cite{mcpherson2001homo,halberstam2016homo,lee2019fair,abbass2018social}. Nevertheless, malicious users can modify topologies to spread false information \cite{roy2021fake}.
Similarly, recommender systems suggest products tailored to our characteristics and history of purchases. However, pursuing the highest accuracy led to the discrimination of minorities in the past \cite{corbett2017fair,obermeyer2019}, despite the law prohibiting unfair treatment based on sensitive traits such as race, religion, and gender. The unfairness arises even if the sensitive attributes are not used explicitly in the learning model. For example, most social networks are homophily-dominant. Nodes in the local neighbourhood belong to the same sensitive class with minimal connections across nodes of differing sensitive attributes. Therefore communities isolate themselves polarizing the opinions expressed within the communities. This effect is also known as the filter bubble problem. The same issue affects the bipartite graphs of users and items used in product recommendations. In \cite{nguyen2014filterbubble}, the authors concluded that recommender systems reduce the exposition of the user to a subset of the items available over time. For example, streaming services may recommend movies from a particular genre to users from a specific gender. Thus, link prediction algorithms have a substantial social impact and can worsen existing biases in the data. However, enforcing the prediction of new links to be fair can mitigate the issue.

Graph neural networks (GNNs) \cite{bronstein2017geometric,bacciu2020gentle,spinelli2021apgcn} provide state-of-the-art link prediction results with an end-to-end learning paradigm. A common approach to improve the fairness of these algorithms requires the introduction of fairness enforcing constraints during a model's training \cite{bose2019cfcge}. Another strategy involves the modification of the graph's topology for post-processing the model's prediction \cite{spinelli2021fairdrop, dai2020fairgnn, loveland2022fairedit}. Along this, the community is studying how to measure the actual fairness introduced in the system by these methods. Link prediction requires a dyadic fairness measure that considers the influence of both sensitive attributes associated with the connection \cite{masrour2020fb}. However, most works on fairness measures focus on independent and identically distributed (i.i.d.) data. A common solution consists in determining new groups defined for the edges. Then, it is possible to measure the level of equity of a new edge added to the graph by applying the known fairness metrics to these new groups.

Since training is the most expensive phase in the modern machine learning pipeline (excluding data harvesting and labelling), we designed a fine-tuning strategy named DEA, where we learn to modify the graph's topology and adapt the parameters of the network to those modifications. A novel covariance-based constraint designed for the link prediction task guides the fine-tuning. We introduce a novel parametrization that allows the new adjacency's optimization in its discrete form. We apply a variation of the Gumbel-max trick \cite{jang2017gumbel}  paired with a small multilayer perceptron that allows us to sample the edges from the original adjacency matrix.

\section{Related Works}
\label{related}
In this section, we focus on the recent contributions to the fair graph representation learning field. Although the extensive and interdisciplinary literature \cite{chiappa2019path,chiappa2020general} on algorithmic bias, the study of fairness in graph representation learning is recent. The surge of interest is due to the state-of-the-art results of graph neural networks (GNNs) in many graph-based tasks.
Some works focused on the node embeddings task to create fair embeddings to use as the input of a downstream link prediction task. Compositional fairness constraints \cite{bose2019cfcge} learn a set of adversarial filters that remove information about particular sensitive attributes. GUIDE \cite{song2022guide} maximize overall individual fairness minimizing at the same time group disparity of individual fairness across different groups. FairWalk \cite{tahleen2019fairwalk} is an adaptation of Node2Vec \cite{grover2016n2v} that aims to increase the fairness of the resulting embeddings. It modifies the transition probability of the random walks at each step, by weighing the neighbourhood of each node, according to their sensitive attributes.
 The recent work of \cite{li2021on} learns a fair adjacency matrix during an end-to-end link prediction task. FairAdj uses a graph variational autoencoder \cite{kipf2016variational} as base architecture and introduces two different optimization processes. One for learning a fair version of the adjacency matrix and one for the link prediction. Similarly, FairDrop \cite{spinelli2021fairdrop} modifies the adjacency during training using biased edge dropout targeting the homophily with respect to the sensitive attribute. However, the biased procedure is non-trainable. FairMod \cite{current2022fairmod}, and FairEdit \cite{loveland2022fairedit} considers debiasing the input graph during training with the addition of artificial nodes and edges and not just the deletion. Except for FairDrop and FairAdj, the other solutions target the task of computing node embeddings or node classification explicitly.
To our knowledge, we are the first to propose a model agnostic fine-tuning strategy to solve the link prediction end-to-end, optimizing both model's utility and fairness protection. Our contribution contains two novelties. From one side, we introduce two covariance-based constraints explicitly to enforce the fairness of the link prediction classification. Secondly, we propose a novel way to parametrize a discrete yet trainable adjacency matrix. The latter aspect is of particular interest to the community to improve the quality of the messages sent across the graph \cite{klicpera2019diff,kazi2022dgm}.
DropEdge \cite{rong2020dropedge} is a dropout mechanism which randomly removes a certain number of edges from an input graph at each training epoch to sparsify the connectivity. In Sparsified Graph Convolutional Network (SGCN) \cite{li2022gs}, the authors first pre-train a GCN to solve a node classification task. Then, a neural network sparsifies the graph by pruning some edges. Finally, they improve the classification performances by training a new GCN on the sparsified graph. Rather than sparsify the topology, another approach consists in rewiring the connections. GraphSage \cite{hamilton2017graphsage} performs a neighbourhood sampling intending to be able to scale to larger graphs. The solution proposed in \cite{klicpera2019diff} alleviates the problem of noisy and often edges in real graphs by combining spectral and spatial techniques. DGM \cite{kazi2022dgm} and IDGL \cite{yu2020idgl} jointly learn the graph structure and graph embedding for a specific task. Finally, taking distance from the message passing framework and using tools from differential geometry, the authors of \cite{topping2022understanding} present a new curvature-based method for graph rewiring. Our solution is closely related to the first approaches sparsifying the topology. However, in future works, we plan to rewire the graphs' topology with the same underlying objective.

\section{Preliminaries}
\label{sec:preliminaries}
\subsection{Graph representation learning}

In this work we will consider an undirected and unweighted graph $\mathcal{G} = (\mathcal{V}, \mathcal{E})$, where $\mathcal{V} = \left\{1, \ldots, n\right\}$ is the set of node indexes, and $\mathcal{E} = \left\{(i, j) \; \vert \; i, j \in \mathcal{V}\right\}$ is the set of arcs (\textit{edges}) connecting pairs of nodes. The meaning of a single node or edge depends on the application. For some tasks, a node $i$ is endowed with a vector $\vc{x}_i \in \mathbb{R}^d$ of features. Each node is also associated with a categorical sensitive attribute $s_i \in S$ (e.g., political preference, ethnicity, gender), which may or may not be part of its features. 
Connectivity in the graph can be summarized by the adjacency matrix $\vc{A} \in \left\{0, 1\right\}^{n\times n}$. This matrix is used to build different types of operators that define the communication protocols across the graph. The vanilla operator is the symmetrically normalized graph Laplacian \cite{kipf2017semi}. A Graph Neural Network $\text{GNN}(\vc X, \vc A)$ can combine node features with the structural information of the graph by solving an end-to-end optimization problem. We will focus on the link prediction task, where the objective is to predict whether two nodes in a network are likely to have a link \cite{nowell2007link}. The output of the GNN consists of a matrix of node embeddings $ \vc H$. Therefore we compute a new $n \times n$ matrix containing a probability score for each possible link in the graph $ \vc{\hat Y} = \text{sigmoid} (\vc H \vc H^T )$. The optimization objective is a binary cross-entropy loss over a subset of positive training edges and negative ones (sampled once).


\subsection{Dyadic group fairness metrics}
\label{subsec:fairness_measures}

Fairness in decision-making is broadly defined as the absence of any advantage or discrimination towards an individual or a group based on their traits \cite{nripsuta2019fdef}. Due to the broadness of the definition, there are several different fairness metrics, each focused on another type of discrimination \cite{mehrabi2019ASO}.
We focus on group fairness metrics measuring if the model's predictions disproportionately benefit or damage people of different groups defined by their sensitive attributes. 

These measures are usually expressed in the context of a binary classification problem. In the notation of the previous section, denote by $Y \in [0,1] $ a binary target variable defined for each node of the graph, and by $\hat Y = f( \vc x)$ a predictor that does not exploit the graph structure. As before, we associate to each $\vc x$ a categorical sensitive attribute $S$. For simplicity's sake, we assume $S$ to be binary, but the following definitions extend easily to the multi-class case.
Two widely used criteria belonging to this group are:
\begin{itemize}
    \item \textit{Demographic Parity} ($DP$) \cite{dwork2012dp}: a classifier satisfies $DP$ if the likelihood of a positive outcome is the same regardless of the value of the sensitive attribute $S$.
    \begin{equation}
        P(\hat Y|S = 1 ) = P(\hat Y|S = 0 )
        \label{eq:demographic_parity}
    \end{equation}
    \item \textit{Equalized Odds} ($EO$) \cite{hardt2016eo}:  a classifier satisfies $EO$ if it has equal rates for true positives and false positives between the two groups defined by the protected attribute $S$.
        \begin{equation}
        P(\hat Y = 1|S = 1 , Y = y ) = P(\hat Y = 1|S = 0, Y = y )
        \end{equation}
\end{itemize}
These definitions trivially extend to cases where the categorical sensitive attribute can have more than two values $|S| > 2$. For the rest of the paper, we will consider this scenario.
The link prediction task's predictive relationship between two nodes should be independent of both sensitive attributes. Therefore, In \cite{masrour2020fb} and \cite{spinelli2021fairdrop}, the authors introduced three dyadic criteria to map the sensitive attributes from the nodes to the edges. The original groups defined by $S$ generate different dyadic subgroups associated with the edges $D$. The dyadic groups can be summarized as follows:

\begin{itemize}
    \item \textbf{Mixed dyadic} $\left(|D| = 2\right)$: the original groups generate two dyadic groups independently from the cardinality of the sensitive attribute. An edge will be in the intra-group if it connects a pair of nodes with the same sensitive attribute. Otherwise, it will be part of the inter-group.
    \item \textbf{Group dyadic} $\left(|D| = |S|\right)$: creates a one-to-one mapping between the dyadic and node-level groups. Each edge is counted twice, once for every sensitive attribute involved. This dyadic definition ensures that the nodes participate in the links' creation regardless of the value of their sensitive attribute.
    \item \textbf{Sub-group dyadic} $\left(|D| = \frac{(|S|+2-1)!}{2!(|S|-1)!} \right)$: enumerates all the possible combinations of sensitive attributes. The fairness criteria protect the balance between all the possible inter-group and intra-group combinations.
\end{itemize}

\section{Drop Edges and Adapt}
\label{sec:method}
\begin{figure*}
    \centering
    \includegraphics[width=1\columnwidth]{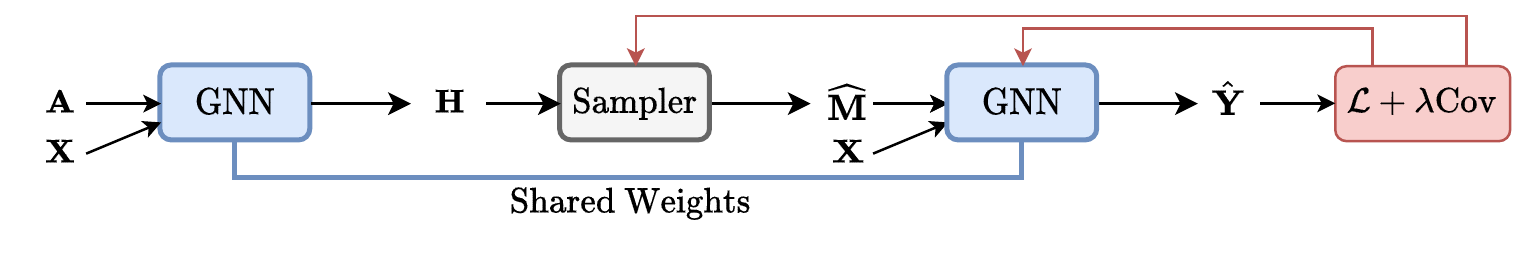}
    \caption{DEA schematics. The pre-trained GNN extracts the node embeddings $\vc H$. The Sampler takes them as input and returns a new, fairness enforcing, discrete adjacency matrix $\widehat{\vc M}$. The new matrix is used as input for a new feedforward step of the GNN. Finally, we update the Sampler and the GNN with a combination of the binary cross-entropy loss and our covariance-based fairness constraint.}
    \label{img:scheme}
\end{figure*}
In this work, we aim to improve the fairness of a trained GNN. In our fine-tuning strategy, we optimize at the same time the model and the adjacency matrix to solve the main task subject to a fairness constraint. To optimize the adjacency matrix, we learn a latent variable for each edge in the original graph with a neural network. The number of the introduced parameters is negligible concerning the size of the input graph, which makes our approach applicable to large-scale datasets. We focus our evaluation on the task of end-to-end link prediction. Therefore we design the constraint accordingly. We show the general framework of our method in Figure \ref{img:scheme}. We aim to fine-tune a trained model with an additional regularization term enforcing fairness by changing the adjacency matrix and adapting the network weights to these modifications. To do so, we introduce a different architecture called Sampler, containing an MLP. The Sampler takes as input the node embeddings produced by the GNN and builds representation for the edges in the graph. Then it outputs a new adjacency which will be used by the GNN to make its predictions. The fine-tuning loss comprises the cross-entropy loss and a fairness constraint that updates the Sampler and the GNN. Below, we introduce each element in a separate section.

\subsection{Sampler}
The Sampler is one of the two key contributions of our proposed approach. We want to sample the edges from the original adjacency matrix to help the GNN produce fairer predictions. At the same time, it has to preserve the discrete nature of the graph during the training process. The Sampler contains an MLP taking as input an edge embedding, defined as the concatenation of two-node embeddings produced by the last layer of the GNN. The output of the MLP is an unnormalized probability vector $\vc z$ where each element is associated with an edge of the graph. To sample the edges, we use the Gumbel-max trick \cite{jang2017gumbel}. It is a method to draw a sample from a categorical distribution, given by its unnormalized (log-)probabilities. The community proposed several extensions of this trick, including a Gumbel-sigmoid \cite{geng2020gs}. We apply this function to the vector $z$:
\begin{equation}
\widetilde m_{(i,j)} = \text{sigmoid}\left(\frac{z_{(i,j)} + G'}{\tau}\right)    
\end{equation}
where $\text{G}'$ is an independent Gumbel noise and $\tau \in (0,\infty)$ is a temperature parameter. As $\tau$ diminishes to zero, a sample from the Gumbel-Sigmoid distribution becomes cold and resembles the one-hot samples. The procedure generates a new vector of soft-noisy weights $\widetilde m_{(i,j)}$ $\forall (i,j) \in \mathcal E$. Finally, we build the new adjacency matrix $\vc{\widehat M}$ where each element is defined as follows:
\begin{equation}
    \widehat m_{(i,j)} =  \begin{cases} 1 & \text { if } \widetilde m_{(i,j)} \geqslant 0.5 \text{ and } (i,j) \in \mathcal E \\ 0 & \text{ otherwise} \end{cases} \,,
\end{equation}
The flowing of the gradient is guaranteed thanks to the use of a straight-through estimator \cite{hinton2012ste}.

\subsection{Constraints}
In \cite{zafar2019constr}, the authors introduce a constraint to design convex boundary-based classifiers free of disparate impact. They use the covariance between the sensitive attribute $s$ and the signed distance from the feature vectors to the decision boundary. Even if this measure is just a proxy for the disparate impact, it led to good empirical results. 
Neural networks, however, are not convex boundary-based classifiers. We cannot apply the constraint in its original formulation. To this end, we propose to exploit the prediction margin instead of the distance from the decision boundary. We recall that the prediction margin for a model parametrized by $\theta$ is defined as follows:
\begin{equation}
    \beta_{\theta}(i,j) = \hat y_{(i,j)} - \delta
\end{equation}
where $\hat y_{(i,j)}$ is the predicted probability for the edge between node $i$ and node $j$. $\delta$ is the threshold to assign the edge to the positive class if $\hat y_{(i,j)} \geqslant\delta $ or otherwise to the negative class.

In our definition of the constraint, we consider the dyadic nature of the link prediction task. The first and most effortless approach consists in building a constraint replicating the mixed dyadic definition. We create a new vector in which we assign to each edge a single value. We let $e=1$ if the nodes at the ends of the edges have the same sensitive attribute and $e=0$ otherwise. The covariance mixed dyadic constraint can be written as:

\begin{equation}
    \text{CovM} = \left \lvert \frac{1}{|\mathcal{E}|} \sum_{(i,j) \in \text{$\mathcal{E}$}}(e_{(i,j)}-\bar e)\beta_\theta(i,j) \right \lvert \leqslant c 
\end{equation}
where $\bar e$ is the mean of the $e$ vector.

We then propose a second version of the constraint mimicking the group dyadic definition to create a more expressive constraint. We create as many vectors as the sensitive attribute $S$ cardinality. The first vector $e^1$ will be associated with the first possible value of the sensitive attribute $S$, denoted as $s^1$ and so on. We then let $e^k_{(i,j)}=1$ if at least $i$ or $j$ has $s^k$ as sensitive attribute. We end up with $|S|$ different $e$ vectors and the same number of covariance constraints. We can minimize the constraint independently by assigning a different threshold $c$  to each one of them or by averaging them together. We can express the latter approach as:

\begin{equation}
    \text{CovG} = \left \lvert \frac{1}{|\mathcal{E}||S|} \sum_{(i,j) \in \mathcal{E}}\sum_{k \in |S|} (e^k_{(i,j)}-\bar e^k)\beta_\theta(i,j) \right \lvert \leqslant c 
\end{equation}

In our evaluation, we opted for the second solution leaving the first approach for future work. In the end, this last approach can be viewed as a one-vs-all fairness constraint where we try to maximize the fairness of all groups at once. 

\subsection{Fine-tuning}

        


    
    
        

Fine-tuning a model has several advantages over training from scratch when one is trying to impose some constraints. First, it is easy to assess the fairness of the prediction of the original model and fine-tuned one. Secondly, it is possible to create a fairer model and obtain a more equitable prediction of the new adjacency without retraining the model. Ideally, we want to optimize the adjacency matrix. However, as it is possible to see in the ablation section, the model suffers drastic changes in its inputs. We found that adapting the model's parameters while learning the adjacency stabilizes the predictive performances meanwhile improving their fairness.
We start with a trained model parameterized by $\theta$ and a threshold value $\delta$ used to assign an edge to the positive or negative class. Next, we sample a negative set of edges for the link prediction loss. For each epoch, we compute the node embeddings. The Sampler takes them as input to output $\widehat{\vc M}$. The network combines this discrete and trainable adjacency with the negative samples for the final feedforward step. Next, we compute the standard cross-entropy for the link prediction task and our covariance-based fairness enforcing constraint. The constraint is balanced with an additional hyperparameter $\lambda$. Finally, we update the GNN and the MLP inside the Sampler.

\section{Experimental section}
\label{sec:evaluation}

\begin{table}[t]
\centering
\scriptsize
\caption{Dataset statistics.}
\begin{tabular}{l|c|c|c|c|c}
Dataset & $S$ & $|S|$ & Features & Nodes & Edges\\
\hline
Citeseer & paper class & 6 & 3703 & 2110 & 3668 \\
Cora-ML  & paper class & 7 & 2879 & 2810 & 7981 \\
PubMed  & paper class & 3 & 500 & 19717 & 44324 \\
DBLP  & continent & 5 & None & 3980 & 6965 \\
FB & gender & 2 & None & 4039 & 88234
\end{tabular}
\label{tab:datasets}
\end{table}

We focus our experiments on measuring the impact of our fine-tuning strategy for enhancing fairness on the link prediction task. We use six fairness metrics (i.e. two for each dyadic group) together with the AUC and accuracy on the main task. In addition, we report the average and standard deviations of ten runs with random data splits.
We monitor the Demographic Parity difference ($\Delta DP$) and the Equality of Odds difference ($\Delta EO$). The first measures the difference between the largest and the lowest group-level selection rate:
\begin{equation}
    \Delta DP = \max_d E[\vc{ \hat Y} | D = d] - \min_d E[\vc{ \hat Y} | D = d]
\end{equation}

The latter report the maximum discrepancy between the true positive rate (TPR) difference and the false positive
rate (FPR) difference between the groups:
\begin{align}  
    \Delta TPR & = \max_d E[\vc{ \hat Y}=1 | D = d, \vc Y=1] \\\nonumber
    & - \min_d E[\vc{ \hat Y}=1 | D = d, \vc Y= 1] \,,
\end{align}
\begin{align}  
    \Delta FPR & = \max_d E[\vc{ \hat Y}=1 | D = d, \vc Y=0] \\\nonumber
    & - \min_d E[\vc{ \hat Y}=1 | D = d, \vc Y=0] \,,
\end{align}
\begin{equation}
\Delta EO = \max( \Delta TPR, \Delta FPR ) \,.
\end{equation}

Our evaluation comprises five datasets. We report their statistics in Table \ref{tab:datasets}. DBLP is a co-authorship network built-in \cite{buyl2020debayes} from the original dataset introduced in \cite{tang2008dbpl}. Nodes represent authors and are connected if they have collaborated at least once. The sensitive attribute is the continent of the author institution without Africa and Antarctica because of their under-representation in the data. Facebook (FB) \cite{leskovec2012data} is a combination of ego-networks introduced in \cite{spinelli2021fairdrop} obtained from a social network. The graph encodes users as nodes with gender as a sensitive attribute and friendships as links. These two datasets do not have feature vectors associated with the nodes. Therefore we used the eigenvectors of the Laplacian matrix as input features.
We included three benchmark citation networks Citeseer, Cora-ML, and PubMed. In these graphs, nodes are articles and have associated a bag-of-words representation of the abstract. Links represent a citation regardless of the direction. We used the category of the article as a sensitive attribute.
We would like to recall that the value of the sensitive attribute arises naturally from the graph topology but is never used directly in the learning pipeline.
We tested our fine-tuning strategy on a GCN \cite{kipf2017semi} and a GAT \cite{velickovic2018graph}. We used an embedding size of 128 for the GCN. The GAT uses an embedding size of 16 with eight attention heads which are concatenated. We used two layers for the citation datasets and four for the two more complex datasets. We chose the threshold for computing the accuracy and the corresponding fairness with a grid search in the interval $[0.4, 0.7]$ for each algorithm.
In our covariance constraints, we set $c=0$ and choose $\lambda$ to balance the regularization term with grid search. The temperature $\tau$ of the Gumbel-sigmoid followed a linear decay from 5 to 1 for each dataset. The MLP in the Sampler has two layers of 128 elements across all experiments. 
We trained the models using Adam optimizer \cite{kingma2014adam} for 100 epochs on every dataset except FB, which required 200 epochs. Our fine-tuning required additional 100 epochs. We compare against competitors designed to enforce the fairness of link prediction tasks. We build upon the experimental evaluation proposed in \cite{spinelli2021fairdrop}. Therefore we include DropEdge and Fairdrop as plain and biased sparsification techniques and FairAdj as a more complex approach. We used two configurations suggested in the original implementation for the latter method. The one with the hyperparameter $T_2=20$ provides a more robust regularization towards fairness with respect to the model trained with $T_2=5$ at the cost of lowering the model's utility.

\subsection{Results}

\begin{table*}[ht]
\tiny
\caption{Link prediction on Citeseer}
\begin{adjustbox}{center}
\begin{tabular}{l|c|c|c c|c c |c c}
 Method & Accuracy $\uparrow$ & AUC $\uparrow$ &  $\Delta DP_{m} \downarrow$ & $\Delta EO_{m}  \downarrow$ & $\Delta DP_{g}  \downarrow$ & $\Delta EO_{g}  \downarrow$ & $\Delta DP_{s}  \downarrow$ & $\Delta EO_{s}  \downarrow$\\
\midrule

GCN & 76.7 $\pm$ 1.3 & 86.7 $\pm$ 1.3 & 42.6 $\pm$ 3.7 & 27.9 $\pm$ 4.7 & 20.6 $\pm$ 4.1 & 22.2 $\pm$ 4.6 & 68.1 $\pm$ 3.7 & 71.4 $\pm$ 9.1 \\
GAT & 76.3 $\pm$ 1.4 & 85.6 $\pm$ 1.9 & 42.4 $\pm$ 2.8 & 26.4 $\pm$ 4.1 & 21.1 $\pm$ 3.8 & 25.4 $\pm$ 5.6 & 71.3 $\pm$ 5.7 & 73.4 $\pm$ 9.9 \\
\midrule
GCN+DropEdge & 78.9 $\pm$ 1.3 & 88.0 $\pm$ 1.3 & 44.9 $\pm$ 2.5 & 27.5 $\pm$ 4.1 & 20.1 $\pm$ 2.9 & 21.6 $\pm$ 5.0 & 71.0 $\pm$ 3.4 & 73.2 $\pm$ 9.5 \\
GAT+DropEdge & 76.3 $\pm$ 0.9 & 85.6 $\pm$ 1.0 & 42.6 $\pm$ 2.5 & 28.4 $\pm$ 5.0 & 22.2 $\pm$ 5.1 & 27.6 $\pm$ 6.3 & 76.7 $\pm$ 3.0 & 77.5 $\pm$ 8.8 \\
\midrule
FairAdj$_{T2=5}$ & 78.5 $\pm$ 2.2 & 86.7 $\pm$ 2.2 & 39.2 $\pm$ 3.2 & 19.0 $\pm$ 3.9 & 17.3 $\pm$ 4.4 & 18.2 $\pm$ 5.8 & 62.6 $\pm$ 4.1 & 47.6 $\pm$ 8.8 \\
FairAdj$_{T2=20}$ & 74.4 $\pm$ 2.5 & 82.5 $\pm$ 2.7 & 31.0 $\pm$ 3.1 & 15.6 $\pm$ 3.0 & \textbf{8.8} $\pm$ 3.2 & 19.7 $\pm$ 6.9 & 56.1 $\pm$ 3.8 & \textbf{43.1} $\pm$ 7.4 \\
\midrule
GCN+FairDrop & \textbf{79.2} $\pm$ 1.4 & \textbf{88.4} $\pm$ 1.4 & 42.6 $\pm$ 2.5 & 26.5 $\pm$ 4.2 & 18.7 $\pm$ 4.0 & 17.6 $\pm$ 5.5 & 67.7 $\pm$ 3.5 & 64.3 $\pm$ 9.5 \\
GAT+FairDrop & 78.2 $\pm$ 1.1 & 87.1 $\pm$ 1.1 & 42.9 $\pm$ 2.2 & 28.3 $\pm$ 4.3 & 22.5 $\pm$ 3.4 & 25.9 $\pm$ 5.2 & 75.3 $\pm$ 3.2 & 73.4 $\pm$ 9.1 \\
\midrule
GCN+DEA+CovM & 79.0 $\pm$ 1.1 & 88.2
$\pm$ 1.1 & \textbf{25.3} $\pm$ 0.5 & 13.3 $\pm$ 0.5 & 11.1 $\pm$ 2.3 & 11.3 $\pm$ 3.1 & 46.7 $\pm$ 3.7 & 48.7
$\pm$ 6.1 \\
GCN+DEA+CovG & 78.8 $\pm$ 1.1 & 88.1 $\pm$ 0.4 & 26.9 $\pm$ 0.9 & 13.4 $\pm$ 2.5 & 9.6 $\pm$ 1.1 & 9.8 $\pm$ 1.3 & \textbf{43.6} $\pm$ 3.3 & 48.9 $\pm$ 5.2 \\

GAT+DEA+CovM & 78.2 $\pm$ 0.4 & 87.9 $\pm$ 0.6 & 37.6 $\pm$ 2.8 & 23.9 $\pm$ 4.1 & 11.6 $\pm$ 3.3 & 12.7 $\pm$ 4.5 & 57.7 $\pm$ 2.3 & 68.4 $\pm$ 7.7 \\
GAT+DEA+CovG & 77.7 $\pm$ 0.4 & 87.5 $\pm$ 0.4 & 38.1 $\pm$ 1.6 & \textbf{9.6} $\pm$ 5.1 & 14.3 $\pm$ 3.8 & \textbf{6.2} $\pm$ 4.4 & 58.8 $\pm$ 2.5 & 57.1 $\pm$ 7.6 \\
\end{tabular}
\end{adjustbox}
\label{tab:citeseer}
\end{table*}

\begin{table*}[ht]
\centering
\tiny
\caption{Link prediction on Cora}
\begin{adjustbox}{center}
\begin{tabular}{l|c|c|c c|c c |c c}
 Method & Accuracy $\uparrow$ & AUC $\uparrow$ &  $\Delta DP_{m} \downarrow$ & $\Delta EO_{m}  \downarrow$ & $\Delta DP_{g}  \downarrow$ & $\Delta EO_{g}  \downarrow$ & $\Delta DP_{s}  \downarrow$ & $\Delta EO_{s}  \downarrow$\\
\midrule
GCN & 81.0 $\pm$ 1.1 & 88.0 $\pm$ 1.0 & 53.5 $\pm$ 2.4 & 34.8 $\pm$ 5.0 & 13.6 $\pm$ 3.2 & 17.7 $\pm$ 4.1 & 88.3 $\pm$ 3.3 & 100.0 $\pm$ 0.0  \\
GAT & 80.2 $\pm$ 1.4 & 88.3 $\pm$ 1.1 & 54.9 $\pm$ 2.9 & 39.6 $\pm$ 4.1 & 12.2 $\pm$ 2.5 & 16.5 $\pm$ 3.4 & 90.9 $\pm$ 3.5 & 100.0 $\pm$ 0.0 \\
\midrule
GCN+DropEdge & \textbf{82.4} $\pm$ 0.9 & \textbf{90.1} $\pm$ 0.7 & 56.4 $\pm$ 2.4 & 36.5 $\pm$ 4.3 & 12.3 $\pm$ 2.6 & 15.4 $\pm$ 3.3 & 90.2 $\pm$ 2.7 & 100.0 $\pm$ 0.0 \\
GAT+DropEdge & 80.5 $\pm$ 1.2 & 88.3 $\pm$ 0.8 & 53.7 $\pm$ 2.5 & 37.1 $\pm$ 3.2 & 18.8 $\pm$ 3.6 & 22.5 $\pm$ 4.2 & 93.6 $\pm$ 2.9 & 100.0 $\pm$ 0.0 \\
\midrule
GCN+FairAdj$_{T2=5}$ & 75.9 $\pm$ 1.6 & 83.0 $\pm$ 2.2 & 40.7 $\pm$ 4.1 & 20.9 $\pm$ 4.3 & 18.4 $\pm$ 2.8 & 31.9 $\pm$ 7.0 & 83.8 $\pm$ 4.9 & \textbf{98.3} $\pm$ 7.2 \\
GAT+FairAdj$_{T2=20}$ & 71.8 $\pm$ 1.6 & 79.0 $\pm$ 1.9 & \textbf{32.3} $\pm$ 2.8 & \textbf{15.8} $\pm$ 4.3 & 23.0 $\pm$ 4.2 & 41.4 $\pm$ 5.9 & 78.3 $\pm$ 6.8 & \textbf{98.3} $\pm$ 7.2 \\
\midrule
GCN+FairDrop & \textbf{82.4} $\pm$ 0.9 & \textbf{90.1} $\pm$ 0.7 & 52.9 $\pm$ 2.5 & 31.0 $\pm$ 4.9 & 11.8 $\pm$ 3.2 & 14.9 $\pm$ 3.7 & 89.4 $\pm$ 3.4 & 100.0 $\pm$ 0.0 \\
GAT+FairDrop & 79.2 $\pm$ 1.2 & 87.8 $\pm$ 1.0 & 48.9 $\pm$ 2.8 & 31.9 $\pm$ 4.3 & 15.3 $\pm$ 3.2 & 18.1 $\pm$ 3.5 & 94.5 $\pm$ 2.0 & 100.0 $\pm$ 0.0 \\
\midrule
GCN+DEA+CovM & 81.5 $\pm$ 0.9 & 89.4 $\pm$ 1.0 & 35.0 $\pm$ 1.4 & 16.5 $\pm$ 4.5 & \textbf{7.8} $\pm$ 1.6 & \textbf{11.9} $\pm$ 3.3 & 67.6 $\pm$ 3.9 & 100.0 $\pm$ 0.0 \\
GCN+DEA+CovG & \textbf{82.4} $\pm$ 0.7 & 89.6 $\pm$ 0.6 & 34.4 $\pm$ 2.4 & \textbf{15.8} $\pm$ 2.4 & 8.9 $\pm$ 2.4 & 12.9 $\pm$ 1.7 & \textbf{65.2} $\pm$ 3.8 & 100.0 $\pm$ 0.0 \\
GAT+DEA+CovM & 81.0 $\pm$ 1.5 & 89.1 $\pm$ 1.4 & 48.1 $\pm$ 2.4 & 29.5 $\pm$ 2.6 & 10.2 $\pm$ 1.1 & 13.3 $\pm$ 2.9 & 85.7 $\pm$ 5.2 & 100.0 $\pm$ 0.0 \\
GAT+DEA+CovG & 81.6 $\pm$ 1.7 & 89.4 $\pm$ 1.2 & 49.8 $\pm$ 0.9 & 32.2 $\pm$ 2.3 & \textbf{7.8} $\pm$ 2.9 & 12.6 $\pm$ 3.0 & 85.6 $\pm$ 3.6 & 100.0 $\pm$ 0.0 \\
\end{tabular}
\end{adjustbox}
\label{tab:cora}
\end{table*}

\begin{table*}[ht]
\centering
\tiny
\caption{Link prediction on PubMed}
\begin{adjustbox}{center}
\begin{tabular}{l|c|c|c c|c c |c c}
 Method & Accuracy $\uparrow$ & AUC $\uparrow$ &  $\Delta DP_{m} \downarrow$ & $\Delta EO_{m}  \downarrow$ & $\Delta DP_{g}  \downarrow$ & $\Delta EO_{g}  \downarrow$ & $\Delta DP_{s}  \downarrow$ & $\Delta EO_{s}  \downarrow$\\
\midrule
GCN & 88.0 $\pm$ 0.4 & 94.5 $\pm$ 0.2 & 43.9 $\pm$ 1.2 & 13.2 $\pm$ 1.4 & 5.0 $\pm$ 1.7 & 4.9 $\pm$ 1.7 & 57.3  $\pm$ 2.0 & 26.2 $\pm$ 3.6 \\
GAT & 86.0 $\pm$ 0.4 & 93.2 $\pm$ 0.3 & 45.2 $\pm$ 0.3 & 18.8 $\pm$ 0.5 & 4.8 $\pm$ 1.8 & 3.5 $\pm$ 0.9 & 57.4 $\pm$ 3.1 & 33.3 $\pm$ 5.1 \\
\midrule
GCN+DropEdge & 88.0 $\pm$ 0.5 & 94.6 $\pm$ 0.3 & 43.7 $\pm$ 1.0 & 12.8 $\pm$ 0.8 & 6.3 $\pm$ 0.7 & 6.0 $\pm$ 1.1 & 57.5 $\pm$ 1.4 & 26.3 $\pm$ 2.3 \\
GAT+DropEdge & 80.0 $\pm$ 1.3 & 88.4 $\pm$ 1.2 & 40.7 $\pm$ 2.7 & 21.2 $\pm$ 2.3 & 4.1 $\pm$ 2.6 & 7.9 $\pm$ 2.8 & 60.8 $\pm$ 3.9 & 50.6 $\pm$ 4.8 \\
\midrule
FairAdj$_{T2=5}$ & 75.5 $\pm$ 2.5 & 84.1 $\pm$ 2.2 & 32.3 $\pm$ 4.7 & 15.9 $\pm$ 4.7 & 7.3 $\pm$ 3.0 & 13.8 $\pm$ 6.2 & 53.4 $\pm$ 9.9 & 43.2 $\pm$ 9.5 \\
FairAdj$_{T2=20}$ & 73.8 $\pm$ 2.4 & 82.1 $\pm$ 2.0 & \textbf{28.9} $\pm$ 4.2 & 14.0 $\pm$ 4.0 & 7.8 $\pm$ 4.0 & 16.5 $\pm$ 6.7 & \textbf{52.5} $\pm$ 9.7 & 43.5 $\pm$ 9.8 \\
\midrule
GCN+FairDrop & 88.4 $\pm$ 0.4 & 94.8 $\pm$ 0.2 & 42.5 $\pm$ 0.5 & \textbf{12.2} $\pm$ 0.7 & 5.6 $\pm$ 1.8 & 5.1 $\pm$ 0.9 & 55.7 $\pm$ 1.5 & 26.6 $\pm$ 2.6 \\
GAT+FairDrop & 81.0 $\pm$ 0.6 & 89.2 $\pm$ 0.6 & 41.7 $\pm$ 1.2 & 22.0 $\pm$ 1.1 & \textbf{3.0} $\pm$ 0.3 & 5.5 $\pm$ 1.6 & 60.7 $\pm$ 2.2 & 49.6 $\pm$ 3.7 \\
\midrule
GCN+DEA+CovM & 88.9 $\pm$ 0.3 & \textbf{95.6} $\pm$ 0.2 & 45.0 $\pm$ 0.8 & \textbf{12.2} $\pm$ 0.7 & 6.0 $\pm$ 0.8 & 4.5 $\pm$ 0.8 & 56.2 $\pm$ 1.8 & \textbf{23.8} $\pm$ 3.9 \\
GCN+DEA+CovG & \textbf{89.0} $\pm$ 0.2 & 95.0 $\pm$ 0.2 & 44.4 $\pm$ 1.0 & 12.3 $\pm$ 0.7 & 5.0 $\pm$ 0.5 & 3.5 $\pm$ 0.6 & 56.6 $\pm$ 1.1 & 24.0 $\pm$ 1.2 \\
GAT+DEA+CovM & 86.1 $\pm$ 0.4 & 93.2 $\pm$ 0.2 & 44.9 $\pm$ 0.9 & 18.2 $\pm$ 0.8 & 4.9 $\pm$ 0.6 & 3.3
$\pm$ 1.1 & 56.5 $\pm$ 2.5 & 30.9 $\pm$ 2.9 \\
GAT+DEA+CovG & 86.2 $\pm$ 0.2 & 93.4 $\pm$ 0.1 & 45.2 $\pm$ 0.6 & 19.1 $\pm$ 0.7 & 4.3 $\pm$ 0.7 & \textbf{2.9} $\pm$ 0.7 & 57.3 $\pm$ 1.8 & 33.2 $\pm$ 2.9 \\
\end{tabular}
\end{adjustbox}
\label{tab:pubmed}
\end{table*}

\begin{table*}[ht]
\tiny
\caption{Link prediction on DBLP}
\begin{adjustbox}{center}
\begin{tabular}{l|c|c|c c|c c |c c}
 Method & Accuracy $\uparrow$ & AUC $\uparrow$ &  $\Delta DP_{m} \downarrow$ & $\Delta EO_{m}  \downarrow$ & $\Delta DP_{g}  \downarrow$ & $\Delta EO_{g}  \downarrow$ & $\Delta DP_{s}  \downarrow$ & $\Delta EO_{s}  \downarrow$\\
\midrule
GCN & 82.4 $\pm$ 0.7 & 86.3 $\pm$ 1.8 & 38.2 $\pm$ 1.0 & 8.3 $\pm$ 3.8 & 27.7 $\pm$ 11.1 & 30.7 $\pm$ 15.7 & 82.3  $\pm$ 9.0 & 96.5 $\pm$ 6.9 \\
GAT & 82.8 $\pm$ 1.0 & 86.4 $\pm$ 1.4 & 37.8 $\pm$ 1.3 & 7.5 $\pm$ 3.6 & 27.5 $\pm$ 9.8 & 32.9 $\pm$ 14.3 & 76.9  $\pm$ 2.2 & 100.0 $\pm$ 0.0 \\
\midrule
GCN+DropEdge & 75.6 $\pm$ 2.2 & 80.5 $\pm$ 1.1 & 30.1 $\pm$ 2.0 & 9.8 $\pm$ 3.4 & 22.3 $\pm$ 4.8 & 25.5 $\pm$ 5.1 & 74.7  $\pm$ 2.0 & 93.8 $\pm$ 8.1 \\
GAT+DropEdge & 77.2 $\pm$ 1.3 & 83.8 $\pm$ 0.7 & 32.6 $\pm$ 2.6 & 7.0 $\pm$ 4.2 & 27.4 $\pm$ 7.9 & 40.7 $\pm$ 19.1 & 74.5  $\pm$ 8.5 & 92.6 $\pm$ 9.1 \\
\midrule
FairAdj$_{T2=5}$ & 57.9 $\pm$ 1.0 & 57.0 $\pm$ 1.0 & \textbf{11.0} $\pm$ 1.4 & 8.8 $\pm$ 1.1 & 18.9 $\pm$ 2.6 & 33.2 $\pm$ 4.2 & \textbf{34.2}  $\pm$ 2.2 & 56.1 $\pm$ 22.4  \\
FairAdj$_{T2=20}$ & 58.5 $\pm$ 0.6 & 57.5 $\pm$ 1.5 & 11.5 $\pm$ 0.8 & 8.5 $\pm$ 1.6 & 18.4 $\pm$ 1.6 & 33.2 $\pm$ 3.4 & 34.3  $\pm$ 1.2 & \textbf{55.9} $\pm$ 22.5  \\
\midrule
GCN+FairDrop & 74.0 $\pm$ 2.0 & 80.8 $\pm$ 1.6 & 27.7 $\pm$ 2.3 & 7.1 $\pm$ 3.4 & 24.3 $\pm$ 9.9 & 29.3 $\pm$ 10.2 & 72.0  $\pm$ 5.4 & 93.8 $\pm$ 8.1\\
GAT+FairDrop & 76.0 $\pm$ 0.6 & 83.3 $\pm$ 0.8 & 30.4 $\pm$ 1.9 & \textbf{5.4} $\pm$ 1.8 & 30.1 $\pm$ 7.7 & 41.5 $\pm$ 14.7 & 71.1  $\pm$ 15.0 & 100.0 $\pm$ 0.0\\
\midrule
GCN+DEA+CovM & 81.4 $\pm$ 1.0 & 86.1 $\pm$ 1.5 & 23.7 $\pm$ 1.5 & 6.9 $\pm$ 2.7 & 20.9 $\pm$ 6.6 & 23.6 $\pm$ 5.5 & 69.2  $\pm$ 9.2 & 96.4 $\pm$ 7.2 \\
GCN+DEA+CovG & 82.4 $\pm$ 1.3 & 85.9 $\pm$ 1.9 & 31.8 $\pm$ 2.8 & 8.0 $\pm$ 3.0 & \textbf{18.3} $\pm$ 5.1 & \textbf{21.1} $\pm$ 4.3 & 73.3  $\pm$ 5.2 & 96.9 $\pm$ 6.2 \\
GAT+DEA+CovM & \textbf{83.6} $\pm$ 1.0 & 86.4 $\pm$ 1.3 & 22.5 $\pm$ 1.0 & 6.8 $\pm$ 2.7 & 20.7 $\pm$ 6.6 & 22.8 $\pm$ 6.1 & 56.5  $\pm$ 8.2 & 96.5 $\pm$ 7.1 \\
GAT+DEA+CovG & \textbf{83.6} $\pm$ 1.2 & \textbf{86.8} $\pm$ 1.5 & 21.6 $\pm$ 1.9 & 6.7 $\pm$ 2.3 & 20.8 $\pm$ 6.5 & 21.7 $\pm$ 5.4 & 76.2  $\pm$ 8.2 & 97.8 $\pm$ 4.4 \\
\end{tabular}
\end{adjustbox}
\label{tab:dblp}
\end{table*}

\begin{table*}[ht]
\tiny
\caption{Link prediction on FB}
\begin{adjustbox}{center}
\begin{tabular}{l|c|c|c c|c c |c c}
 Method & Accuracy $\uparrow$ & AUC $\uparrow$ &  $\Delta DP_{m} \downarrow$ & $\Delta EO_{m}  \downarrow$ & $\Delta DP_{g}  \downarrow$ & $\Delta EO_{g}  \downarrow$ & $\Delta DP_{s}  \downarrow$ & $\Delta EO_{s}  \downarrow$\\
\midrule
GCN & 82.3 $\pm$ 1.4 & 90.8 $\pm$ 1.1 & 0.7 $\pm$ 0.5 & 2.8 $\pm$ 0.7 & 3.5 $\pm$ 0.7 & 3.4 $\pm$ 1.2 & 7.5  $\pm$ 1.2 & 5.6 $\pm$ 2.0 \\
GAT & 80.2 $\pm$ 2.7 & 86.3 $\pm$ 2.4 & 0.8 $\pm$ 0.6 & 1.8 $\pm$ 0.9 & 3.3 $\pm$ 1.3 & 2.5 $\pm$ 1.5 & 7.2  $\pm$ 2.5 & 4.3 $\pm$ 2.5 \\

\midrule
GCN+DropEdge & 77.9 $\pm$ 1.5 & 87.7 $\pm$ 0.9 & 0.9 $\pm$ 0.1 & 4.9 $\pm$ 0.4 & 5.5 $\pm$ 0.5 & 6.9 $\pm$ 0.8 & 11.0  $\pm$ 1.0 & 11.7 $\pm$ 1.5\\
GAT+DropEdge & 71.7 $\pm$ 4.6 & 83.9 $\pm$ 1.5 & 1.1 $\pm$ 0.2 & 1.4 $\pm$ 1.4 & 2.3 $\pm$ 1.3 & 2.4 $\pm$ 2.6 & \textbf{5.3}  $\pm$ 2.0 & 4.4 $\pm$ 4.5 \\
\midrule
GCN+FairDrop & 77.4 $\pm$ 1.9 & 87.7 $\pm$ 1.0 & \textbf{0.6} $\pm$ 0.3 & 4.3 $\pm$ 0.2 & 4.9 $\pm$ 0.2 & 5.9 $\pm$ 0.7 & 10.0  $\pm$ 0.5 & 10.1 $\pm$ 1.2\\
GAT+FairDrop & 75.1 $\pm$ 2.1 & 83.7 $\pm$ 1.2 & 1.3 $\pm$ 0.7 & 1.8 $\pm$ 1.5 & 2.3 $\pm$ 1.4 & 2.5 $\pm$ 2.4 & 5.6  $\pm$ 2.4 & 4.7 $\pm$ 4.0 \\
\midrule
GCN+DEA+CovM & \textbf{82.9} $\pm$ 1.2 & \textbf{93.5} $\pm$ 0.6 & 1.6 $\pm$ 0.3 & 1.9 $\pm$ 0.7 & 2.6 $\pm$ 0.3 & 1.6 $\pm$ 0.6 & 6.1  $\pm$ 0.7 & 2.9 $\pm$ 0.9 \\
GCN+DEA+CovG & \textbf{82.9} $\pm$ 1.2 & 93.2 $\pm$ 0.9 & 1.9 $\pm$ 0.3 & 1.6 $\pm$ 0.3 & \textbf{2.1} $\pm$ 0.4 & 1.4 $\pm$ 0.3 & 5.5  $\pm$ 0.6 & 2.5 $\pm$ 0.4 \\
GAT+DEA+CovM & 82.8 $\pm$ 2.3 & 90.9 $\pm$ 2.0 & 1.1 $\pm$ 0.6 & 1.5 $\pm$ 0.9 & 2.9 $\pm$ 0.6 & 1.4 $\pm$ 0.8 & 6.4  $\pm$ 1.2 & 2.7 $\pm$ 1.2 \\
GAT+DEA+CovG & 82.2 $\pm$ 2.1 & 89.5 $\pm$ 2.1 & 1.3 $\pm$ 0.4 & \textbf{1.0} $\pm$ 0.5 & 2.5 $\pm$ 0.8 & \textbf{1.0} $\pm$ 0.6 & 5.8  $\pm$ 1.8 & \textbf{2.2} $\pm$ 1.0 \\
\end{tabular}
\end{adjustbox}
\label{tab:fb}
\end{table*}

\begin{figure*}
    \centering
    \subfigure[$\vc z$]{\includegraphics[width=0.40\columnwidth]{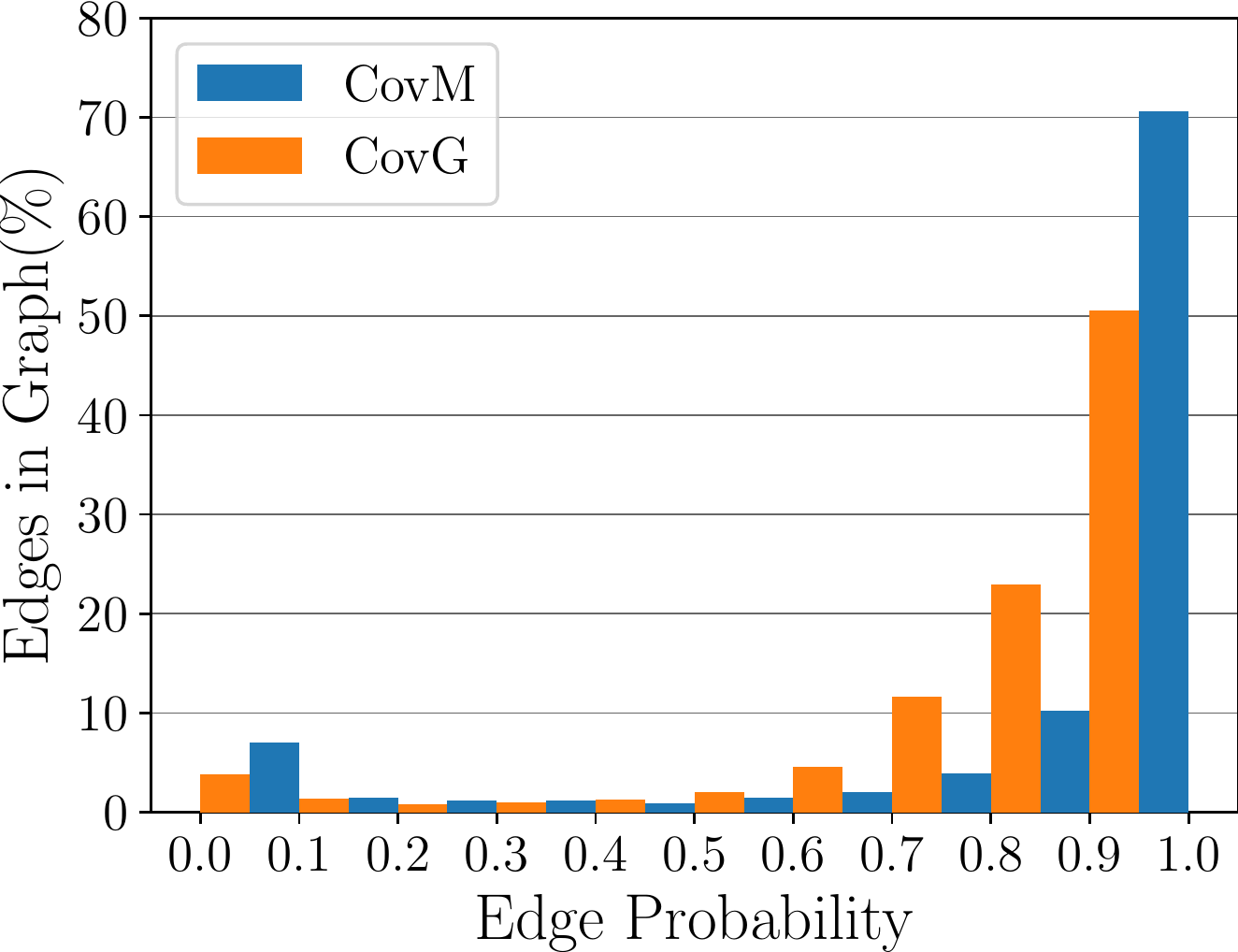}}
    \subfigure[$\widetilde{\vc m}$]{\includegraphics[width=0.40\columnwidth]{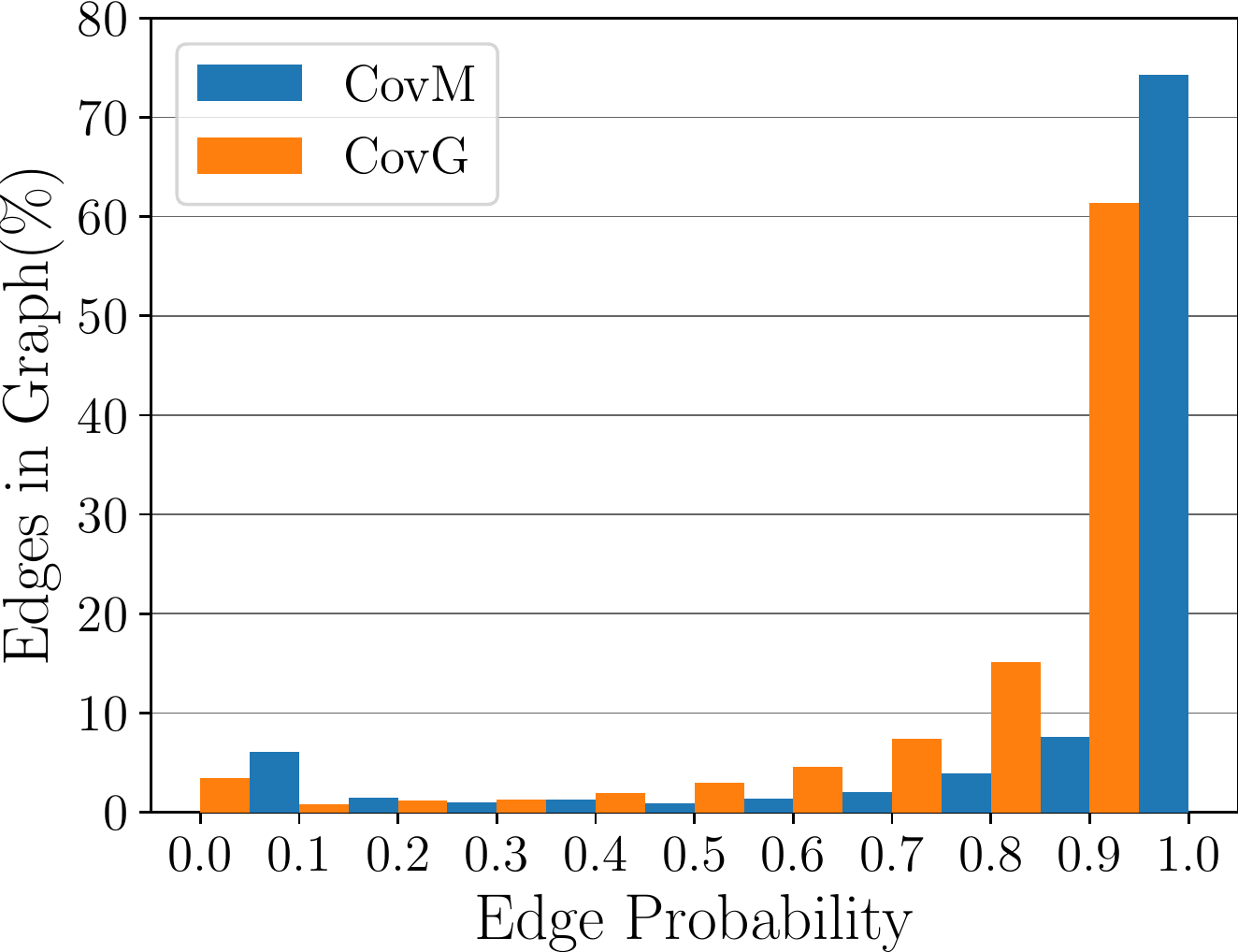}}
    \subfigure[$\widehat{\vc M}$]{\includegraphics[width=0.40\columnwidth]{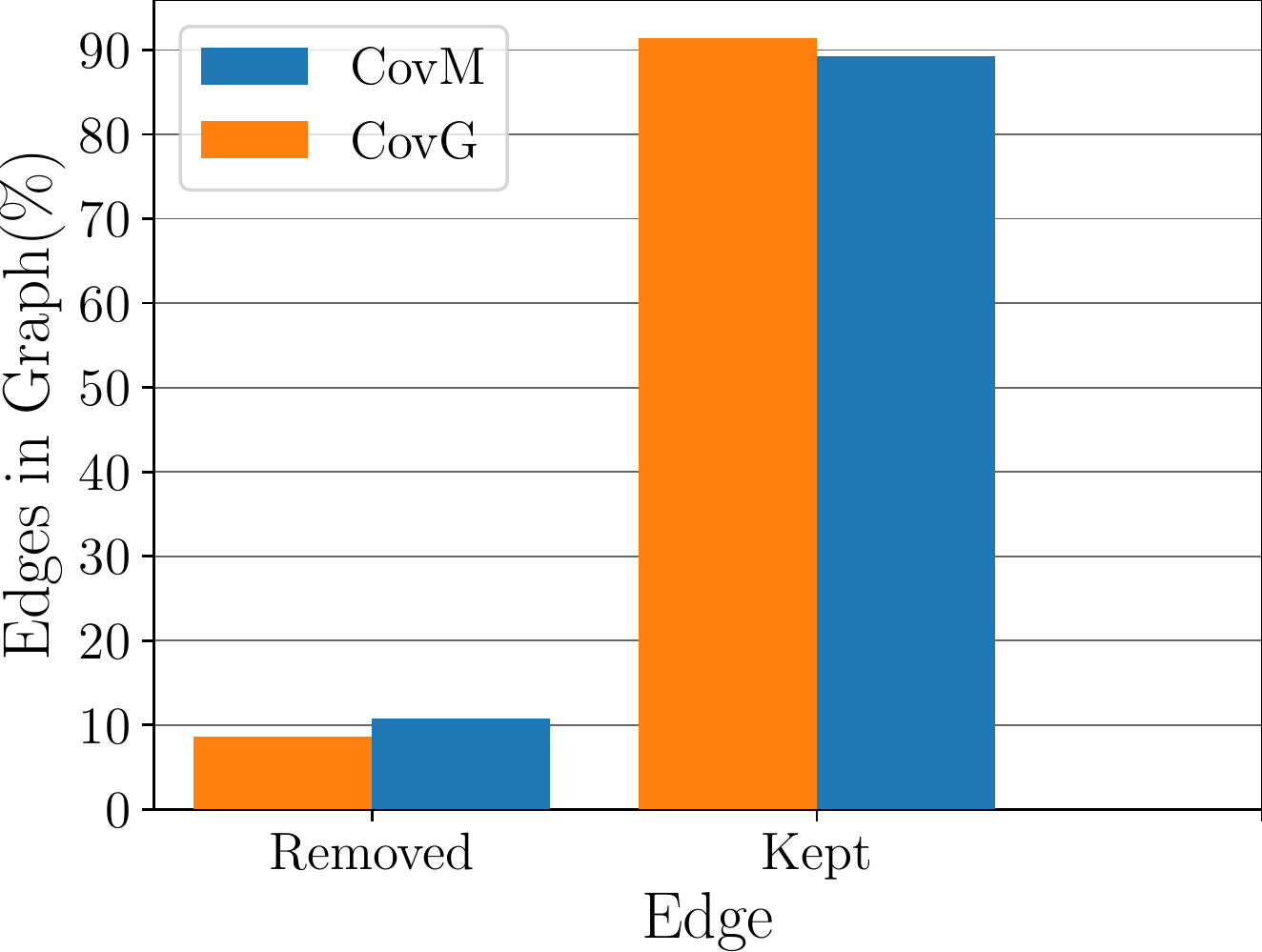}}
    \caption{Edge distribution at different stages of our pipeline. In blue, we depict the results obtained using CovM constraint; in orange, the ones with CovG. Figure (a) shows the distribution $\vc z$ learnt by the MLP inside our Sampler. Figure (b) shows the approximation after the Gumbel sigmoid trick $\widetilde{\vc m}$. Finally, Figure (c) shows the number of edges removed and kept in the new fairness-enforcing adjacency matrix $\widehat{\vc M}$ thresholding the values in $\widetilde{\vc m}$ at $0.5$.}
    \label{img:citeseer}
\end{figure*}

We present the results in Tables \ref{tab:citeseer}, \ref{tab:cora}, \ref{tab:pubmed}, \ref{tab:dblp} and \ref{tab:fb}. DEA with CovM and CovG improves the accuracy on larger datasets and provides state-of-the-art protection against unfairness. CovM shows better fairness metrics in the mixed dyadic group. CovG has a slight advantage on the group dyadic definition. Since the constraint definition closely resembles the group definitions, this is not surprising. CovG has a slight general advantage, probably due to the additional expressiveness of the constraint. On smaller datasets, Tables \ref{tab:citeseer}, \ref{tab:cora} and \ref{tab:pubmed}, DEA provides slightly better protection than FairAdj. However, the latter loses in accuracy and AUC with severe losses in Table \ref{tab:pubmed} where the drop in accuracy provided by FairAdj is about $15\%$ of accuracy and $10\%$ of AUC. FairAdj fails to solve the link prediction task on complex datasets like DBLP (Tab. \ref{tab:dblp}) and FB (Tab. \ref{tab:fb}).
In the end, DEA  removes around $10\%$ of the edges, considerably less than DropEdge and FairDrop. 
In Figure \ref{img:citeseer}, we show the intermediate steps resulting in the final version of the fair adjacency matrix $\widehat{\vc M}$. There is little difference between the actual edge distribution $\vc Z$ and its noisy version after the Gumbel-sigmoid trick $\vc{\widetilde M}$. Also, it is possible to see that CovM is more peaked at the extreme values. Finally, Figure \ref{img:citeseer}(c) shows the number of edges removed from the original adjacency matrix $ \widehat{\vc M}$ to obtain a fairer link prediction.

\subsection{Ablation}
\label{sec:ablation}

\begin{table}[t]
\centering
\tiny
\caption{Results obtained training a GCN on Citeseer with and without the Sampler optimizing the adjacency matrix.}

\begin{tabular}{l| c c | c c}
 Method & Acc Sampler  & AUC Sampler & Acc w/o  & AUC w/o \\
\midrule
Citeseer & \textbf{78.3} $\pm$ 0.5 & \textbf{88.6} $\pm$ 0.3 & 76.7 $\pm$ 1.3 & 86.7 $\pm$ 1.3 \\
\midrule
Cora & \textbf{82.1} $\pm$ 1.1 & \textbf{89.8} $\pm$ 1.0 & 81.0 $\pm$ 1.1 & 88.0 $\pm$ 1.0\\
\midrule
Pubmed & \textbf{88.8} $\pm$ 0.5 & \textbf{95.0} $\pm$ 0.3 & 88.0 $\pm$ 0.4 & 94.5 $\pm$ 0.2\\
\midrule
DBLP & \textbf{83.2} $\pm$ 1.0 & \textbf{86.9} $\pm$ 0.9 & 82.4 $\pm$ 0.7 & 86.3 $\pm$ 1.8 \\
\midrule
FB & \textbf{82.6} $\pm$ 0.9 & 90.7 $\pm$ 0.6 & 82.3 $\pm$ 1.4 & \textbf{90.8} $\pm$ 1.1 \\
\end{tabular}

\label{tab:ablation1}
\end{table}

\begin{table*}[ht]
\centering
\tiny
\caption{Ablation study on Citeseer with CovM}
\label{tab:ablation2}
\begin{adjustbox}{center}
\begin{tabular}{l|c|c|c c|c c |c c}
 Method & Accuracy $\uparrow$ & AUC $\uparrow$ &  $\Delta DP_{m} \downarrow$ & $\Delta EO_{m}  \downarrow$ & $\Delta DP_{g}  \downarrow$ & $\Delta EO_{g}  \downarrow$ & $\Delta DP_{s}  \downarrow$ & $\Delta EO_{s}  \downarrow$\\
\midrule
GCN+CovM & \textbf{79.0} $\pm$ 1.1 & 88.4
$\pm$ 1.1 & \textbf{25.3} $\pm$ 0.5 & \textbf{13.3} $\pm$ 0.5 & \textbf{11.1} $\pm$ 2.3 & \textbf{11.3} $\pm$ 3.1 & \textbf{46.7} $\pm$ 3.7 & \textbf{48.7}
$\pm$ 6.1 \\
\midrule
Training & 78.8 $\pm$ 1.0 & 88.2
$\pm$ 0.9 & 45.5 $\pm$ 2.9 & 28.9 $\pm$ 7.2 & 21.2 $\pm$ 4.4 & 23.8 $\pm$ 5.3 & 73.0 $\pm$ 4.0 & 59.3
$\pm$ 9.4 \\
\midrule
Training w X & 78.4 $\pm$ 1.5 & 88.1
$\pm$ 1.2 & 44.3 $\pm$ 3.6 & 28.0 $\pm$ 6.9 & 21.4 $\pm$ 5.1 & 25.1 $\pm$ 6.8 & 73.1 $\pm$ 1.7 & 59.3 $\pm$ 6.5 \\
\midrule
w/o Sampler & 78.8 $\pm$ 0.7 & 87.8
$\pm$ 0.5 & 30.9 $\pm$ 1.9 & 14.9 $\pm$ 4.1 & 13.4 $\pm$ 2.4 & 14.2 $\pm$ 2.4 & 48.5 $\pm$ 2.5 & 58.9
$\pm$ 8.6 \\
\midrule
w/o CovM & 78.4 $\pm$ 0.6 & 88.0
$\pm$ 0.5 & 29.3 $\pm$ 0.6 & 18.1 $\pm$ 4.6 & 11.9 $\pm$ 1.9 & 13.3 $\pm$ 2.6 & 51.0 $\pm$ 3.9 & 63.8
$\pm$ 9.8\\
\midrule
w/o Sampler \& CovM & 78.9 $\pm$ 0.9 & \textbf{88.6}
$\pm$ 0.8 & 45.2 $\pm$ 2.1 & 29.1 $\pm$ 4.1 & 20.5 $\pm$ 3.7 & 23.3 $\pm$ 7.4 & 73.6 $\pm$ 4.2 & 61.8
$\pm$ 6.6 \\

\end{tabular}
\end{adjustbox}
\end{table*}

In this section, we perform an in-depth ablation study to shed more light on the effect of these additional epochs and each component of the framework. In the first experiment, we train for the same number of epochs as a standard GCN and one paired with the Sampler, learning a new adjacency matrix with the sole objective of maximizing accuracy. Then, we optimize the adjacency matrix and the model's parameters to solve the main task without additional fairness constraints. Finally, in Table \ref{tab:ablation1}, we show that those modifications to the adjacency matrix improve the link prediction performances.
In the second, we focus on the various components of our architecture on the Citeseer dataset. Results are visible in Table \ref{tab:ablation2}. We disable each time a different component of our framework. We train everything from scratch instead of fine-tuning a model in the second and third rows. In ``Training w X", we feed to the Sampler the concatenation of the feature vectors associated with the nodes instead of the node embeddings generated by the GNN. We then proceed to fine-tune the model by disabling some components. In ``w/o Sampler", we keep the covariance constraint but remove the learning of the adjacency matrix. In ``w/o CovM", we do the opposite. Finally, we fine-tune the model without any modification. The latter solution has comparable performances in terms of accuracy, but it has significantly worst fairness metrics. Training from scratch has similar results. Fine-tuning with the covariance constraint or the Sampler improves the fairness, but we obtain the best results when both are active.

\section{Conclusions}
\label{sec:conclusion}

We introduced DEA, a novel approach to improve the fairness of a GNN solving a link prediction task. In our fine-tuning strategy, we learn to modify the graph's topology and adapt the parameters of the network to those modifications. A module called Sampler learns to drop edges from the original adjacency matrix. We exploit a Gumbel-sigmoid to sample a new discrete and fair adjacency. At the same time, the GNN uses this new matrix for fine-tuning. We guide both optimization processes with an additional regularization term shaped as a covariance-based constraint. We provided two different formulations, the first acting on the inter and intra connections between the groups defined by the sensitive attribute. In the second modelling, each value of the sensitive attribute is in the one-vs-the-rest paradigm. We performed an extensive experimental evaluation where we demonstrated that our fine-tuning strategy provides state-of-the-art protection against unfairness meanwhile improving the model's utility on the original task. Finally, we performed an ablation study on the contribution of each component of our pipeline. In future, we would like to learn to add new connections instead of just dropping them from the original adjacency matrix.

\bibliography{biblio}
\bibliographystyle{apalike}
\end{document}